\documentclass[conference]{IEEEtran}

\usepackage[utf8]{inputenc}

\usepackage{booktabs}
\usepackage[font=footnotesize]{caption} 
\usepackage{cite}
\usepackage{float}
\usepackage[acronym]{glossaries}
\usepackage{graphicx}
\usepackage{lipsum}
\usepackage{subcaption}
\usepackage{xcolor}

\graphicspath{Figures}

\newacronym {aer}   {AER}   {address-event representation}
\newacronym {adex}  {AdEx}  {Adaptive Exponential Integrate-and-Fire}
\newacronym {ai}    {AI}    {artificial intelligence}
\newacronym {ann}   {ANN}   {artificial neural network}
\newacronym {ap}    {AP}    {action potential}
\newacronym {api}   {API}   {application programming interface}

\newacronym {cam}   {CAM}   {content-addressable memory}
\newacronym {cmos}  {CMOS}  {complementary metal–oxide–semiconductor}
\newacronym {cnn}   {CNN}   {convolutional neural network}
\newacronym {cps}   {CPS}   {cyber-physical system}
\newacronym {cpu}   {CPU}   {central processing unit}

\newacronym {dl}    {DL}    {deep learning}
\newacronym {dnn}   {DNN}   {deep neural network}
\newacronym {dpi}   {DPI}   {differential-pair integrator}
\newacronym {dstc}  {dSTC}  {Disynaptic Spatiotemporal Correlator}
\newacronym {dynap} {DYNAP} {Dynamic Neuromorphic Asynchronous Processor}

\newacronym {e-i}   {E--I}  {excitatory--inhibitory}
\newacronym {epsc}  {EPSC}  {excitatory postsynaptic current}
\newacronym {epsp}  {EPSP}  {excitatory postsynaptic potential}
\newacronym {exc}   {Exc.}  {excitatory}

\newacronym {fac}   {Fac.}  {facilitatory}
\newacronym {fdhm}  {FDHM}  {full duration at half maximum/minimum}
\newacronym {fnr}   {FNR}   {false-negative rate}
\newacronym {fpga}  {FPGA}  {field-programmable gate array}
\newacronym {fpr}   {FPR}   {false-positive rate}

\newacronym {gals}  {GALS}  {globally asynchronous locally synchronous}
\newacronym {gpu}   {GPU}   {graphics processing unit}

\newacronym {hil}   {HIL}   {hardware-in-the-loop}

\newacronym {ict}   {ICT}   {information and communications technology}
\newacronym {inh}   {Inh.}  {inhibitory}
\newacronym {iot}   {IoT}   {Internet of things}
\newacronym {ipsp}  {IPSP}  {inhibitory postsynaptic potential}
\newacronym {ipi}   {IPI}   {interpulse interval}
\newacronym {isi}   {ISI}   {interspike interval}

\newacronym {ltd}   {LTD}   {long-term depression}
\newacronym {ltp}   {LTP}   {long-term potentiation}

\newacronym {mos}       {MOS}       {metal oxide semiconductor}
\newacronym {mosfet}    {MOSFET}    {metal–oxide–semiconductor field-effect transistor}

\newacronym {nce}   {NCE}   {neuromorphic computing and engineering}

\newacronym {pir}   {PIR}   {postinhibitory rebound}
\newacronym {psp}   {PSP}   {postsynaptic potential}
\newacronym {pvt}   {PVT}   {power, voltage, and temperature}

\newacronym {stc}   {STC}   {Spatiotemporal Correlator}
\newacronym {snn}   {SNN}   {spiking neural network}
\newacronym {sram}  {SRAM}  {static random-access memory}
\newacronym {stdp}  {STDP}  {spike-timing-dependent plasticity}

\newacronym {tde}   {TDE}   {time-difference encoder}
\newacronym {tnr}   {TNR}   {true-negative rate}
\newacronym {tpr}   {TPR}   {true-positive rate}
\newacronym {trig}  {Trig.} {trigger}

\newacronym {vlsi}  {VLSI}  {very-large-scale integration}

\begin{document}

\title{%
    A Comparison of Temporal Encoders for\\
    Neuromorphic Keyword Spotting with Few Neurons
}

\author{%
    \IEEEauthorblockN{%
    Mattias~Nilsson%
    \IEEEauthorrefmark{1},
    Ton~Juny~Pina%
    \IEEEauthorrefmark{2}%
    \IEEEauthorrefmark{3},
    Lyes~Khacef%
    \IEEEauthorrefmark{2}%
    \IEEEauthorrefmark{3}, 
    Foteini~Liwicki%
    \IEEEauthorrefmark{1},
    Elisabetta~Chicca%
    \IEEEauthorrefmark{2}%
    \IEEEauthorrefmark{3},
    and Fredrik~Sandin%
    \IEEEauthorrefmark{1}
}
\IEEEauthorblockA{%
    \IEEEauthorrefmark{1}%
    \itshape
    Embedded Intelligent Systems Lab (EISLAB),
    Luleå University of Technology,
    971\ 87 Luleå,
    Sweden\\
    \upshape
    mattias.k.nilsson@proton.me
}
\IEEEauthorblockA{%
    \IEEEauthorrefmark{2}%
    \itshape
    Bio-Inspired Circuits and Systems (BICS) Lab,
    Zernike Institute for Advanced Materials,
    University of Groningen,\\
    9747 AG Groningen,
    The Netherlands\\
}
\IEEEauthorblockA{%
    \IEEEauthorrefmark{3}%
    \itshape
    Groningen Cognitive Systems and Materials Center (CogniGron),
    University of Groningen,
    9747 AG Groningen,
    The Netherlands\\
    \upshape
    t.juny.pina@rug.nl
}
    %
}

\maketitle

\begin{abstract}
With the expansion of AI-powered virtual assistants, there is a need for low-power keyword spotting systems providing a “wake-up” mechanism for subsequent computationally expensive speech recognition.
One promising approach is the use of neuromorphic sensors and spiking neural networks (SNNs) implemented in neuromorphic processors for sparse event-driven sensing.
However, this requires resource-efficient SNN mechanisms for temporal encoding, which need to consider that these systems process information in a streaming manner, with physical time being an intrinsic property of their operation.
In this work, two candidate neurocomputational elements for temporal encoding and feature extraction in SNNs described in recent literature---the spiking time-difference encoder (TDE) and disynaptic excitatory--inhibitory (E--I) elements---are comparatively investigated in a keyword-spotting task on formants computed from spoken digits in the TIDIGITS dataset.
%
%
While both encoders improve performance over direct classification of the formant features in the training data, enabling a complete binary classification with a logistic regression model, they show no clear improvements on the test set.
Resource-efficient keyword spotting applications may benefit from the use of these encoders, but further work on methods for learning the time constants and weights is required to investigate their full potential.
%
\end{abstract}


\section{Introduction}
\label{sec:introduction}


Speech recognition is becoming an increasingly important computational task, as the number of technological devices using vocal interaction---``voice assistants"---are becoming more common.
This is a computationally expensive task, due to the model complexity required to recognize thousands of different spoken words.
Keyword spotting, in contrast, is limited to the recognition of one or a few specific words or phrases, and thus has a significantly lower demand for computational complexity and resources.
Therefore, keyword spotting is used to ``wake up" vocal-interaction devices, for them to subsequently activate more resource-intensive speech-recognition computation.


%
%
The new generation of brain-inspired neuromorphic processors \cite{roy2019towards} paves the way towards the development of ultra-low-power models for keyword spotting.
These devices comprise in-memory computational elements with massive parallelism and event-driven operation, which allow spatiotemporally sparse, efficient real-time processing for brain-like computation \cite{davies2021advancing}.
Due to the need for low-power keyword spotting, the application of neuromorphic computing for this task has been a subject of recent research \cite{coath2014robust, blouw2019benchmarking, weidel2021wavesense, yin2022attentive}.
%
%
However, these models rely on various methods for temporal encoding.
Some of these methods, such as buffering, transmission delays, and finely tuned time constants, are not optimally suited for efficient real-time implementation in mixed-signal neuromorphic hardware.


Temporal encoding is an important aspect of efficient neuromorphic real-time processing of time-varying patterns \cite{indiveri2019importance}.
Physical time is an intrinsic part of the operations of \glspl{snn}, and this can be used in various ways to efficiently represent relations in processed data \cite{auge2021survey}.
Furthermore, for the creation of resource-efficient neuromorphic implementations, such as for keyword spotting, it is of interest to minimize the number of neurons and spikes that are required for successful classification.


Here, we compare the use of two different \gls{snn} mechanisms for temporal encoding in a keyword spotting task on the TIDIGITS dataset \cite{leonard1993tidigits}.
The investigated encoders are the \gls{tde} \cite{milde2018elementary} and disynaptic \gls{e-i} elements \cite{sandin2020delays}, which both target low-power implementation in neuromorphic hardware.
The \gls{tde} encodes time differences between spikes on pairs of input channels, while the \gls{e-i} elements offer an alternative to delays that leverage the heterogeneity of analog hardware for representational variance.
We used both encoders to process the same input data---formants of spoken digits---and comparatively evaluate their respective effect on subsequent linear classification.


We find that both encoders, respectively, enable a complete binary classification of the training set, compared to the 79\%--96\% accuracy of direct linear classification of the formant input data.
On the test data, however, there is no clear performance improvement by using either encoder.
%
%
We conclude that both investigated approaches may have potential for resource-efficient keyword spotting applications, while there is a seeming need to increase the neural heterogeneity and to develop or implement training protocols \cite{perez2021neural, bouanane2022impact} in both cases.



\section{Methods}
\label{sec:methods}

\begin{figure}[tb]
    \centering
    \includegraphics[width=\columnwidth]{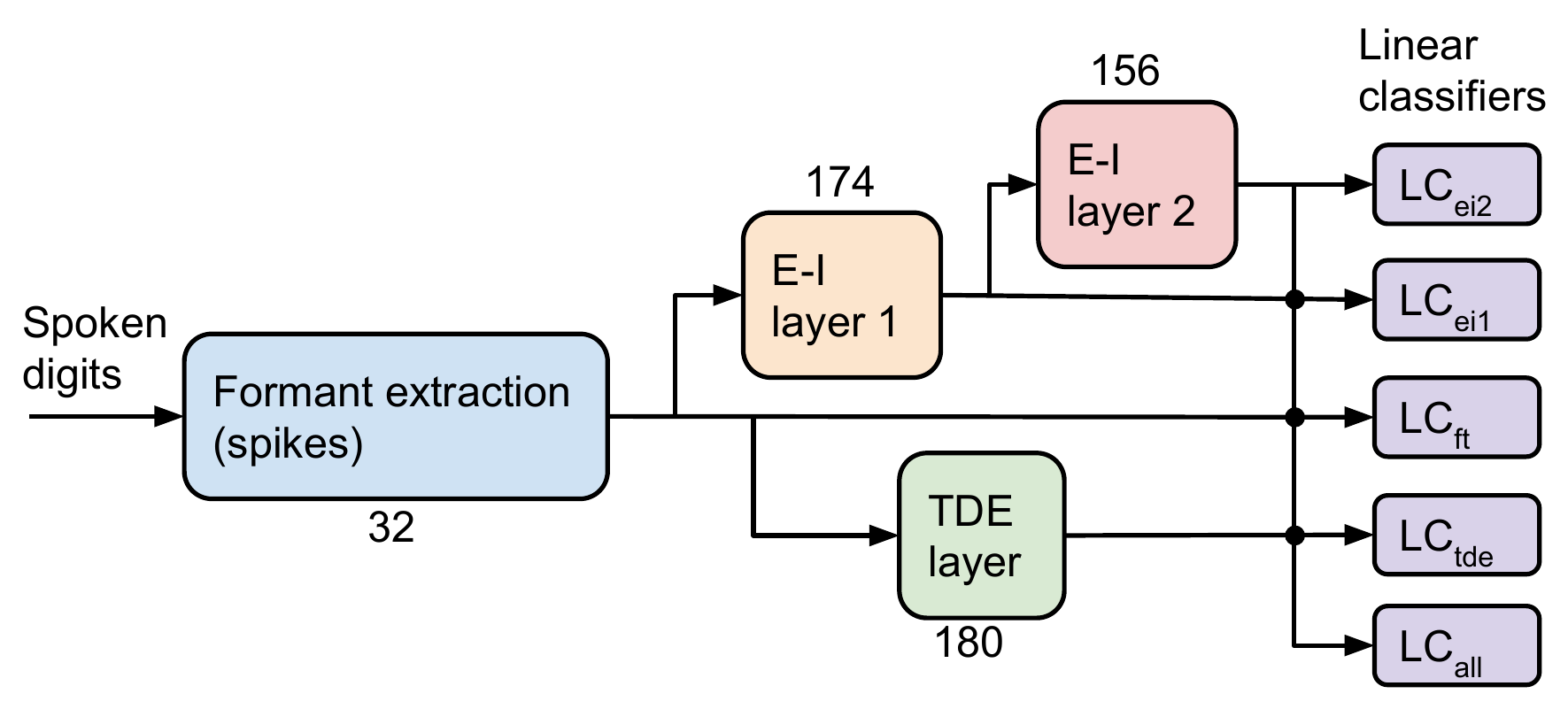}
    \caption{%
        \textbf{Spiking neural networks for keyword spotting with different temporal encoding layers.}
        The linear classifiers (LCs) employ a logistic regression model, which is fitted using $\ell_2$ regularization.
    }
    \label{fig:network}
\end{figure}

A schematic representation of the investigated system is presented in \textbf{Fig.~\ref{fig:network}}.
The input data is processed in parallel through a layer of TDE neurons \cite{milde2018elementary}, see Section~\ref{sec:tde}, and two sequential layers of neurons with disynaptic E--I elements \cite{sandin2020delays, nilsson2022spatiotemporal}, see Section~\ref{sec:ei_elements}.
The system comparatively assesses the improvement in linear classification performance provided by each neurosynaptic structure on a keyword spotting task, see Section~\ref{sec:task}.
The linear classifiers employ a logistic regression model, which was fitted using the LIBLINEAR solver in scikit-learn with $\ell_2$ regularization.

%

\subsection{Task and Dataset}

\label{sec:task}
The keyword-spotting task addressed in this work was set up in the form of \textit{one-vs.-rest} binary classification of spoken digits from the TIDIGITS dataset \cite{leonard1993tidigits}.
Specifically, we used all single-digit utterances (``oh" and ``zero" to ``nine") for all adult speakers---111 men and 114 women, respectively---as was done in \cite{wu2018biologically}.
We used the default training--test split of the dataset, which partitions each speaker group---``men" and ``women"---into test and training subsets containing 224 and 226 samples per digit, respectively.
We balanced the dataset, while using all samples of the keyword digit, by dividing the number of samples for each non-keyword digit by the total number of non-keyword digits (10) and rounding to the nearest integer value.
Thus, for each keyword in the keyword spotting task, the dataset consisted of 224 vs.\ 10~$\times$~22 training samples, and 226 vs.\ 10~$\times$~23 test samples, respectively.

\begin{figure*}[!htb]
    \centering
    \begin{subfigure}[b]{0.49\textwidth}
        \centering
        \includegraphics[width=\textwidth]{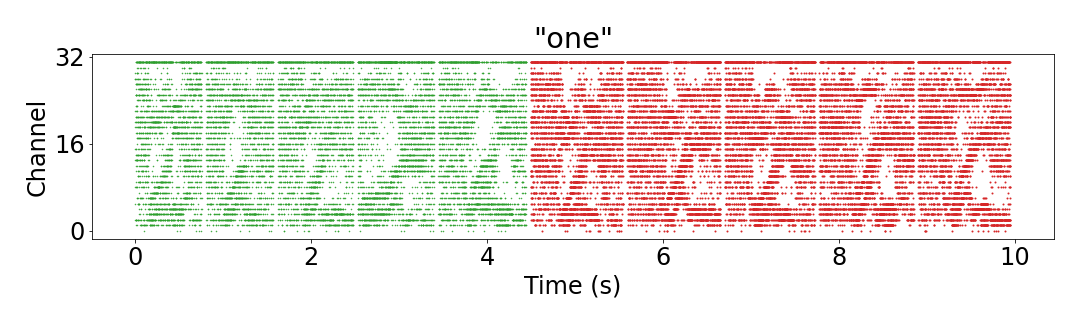}
     \end{subfigure}
     \begin{subfigure}[b]{0.49\textwidth}
        \centering
        \includegraphics[width=\textwidth]{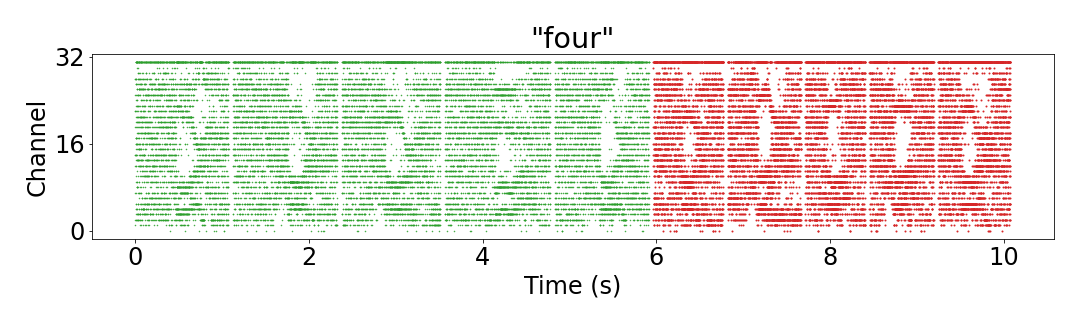}
    \end{subfigure}
    \\
     \begin{subfigure}[b]{0.49\textwidth}
        \centering
        \includegraphics[width=\textwidth]{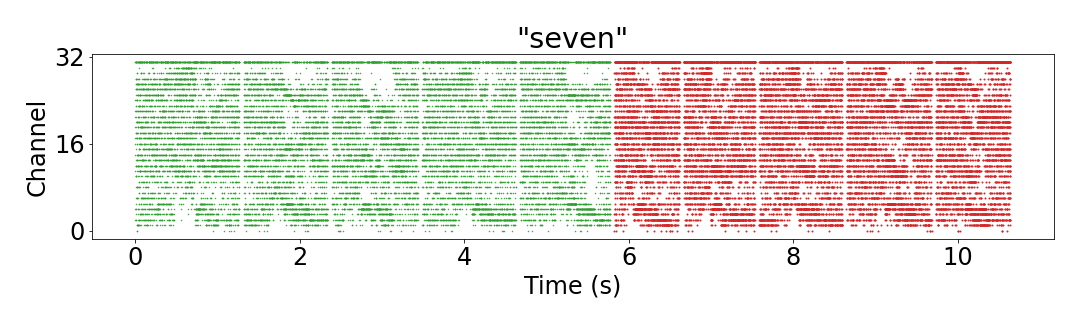}
    \end{subfigure}
    \begin{subfigure}[b]{0.49\textwidth}
        \centering
        \includegraphics[width=\textwidth]{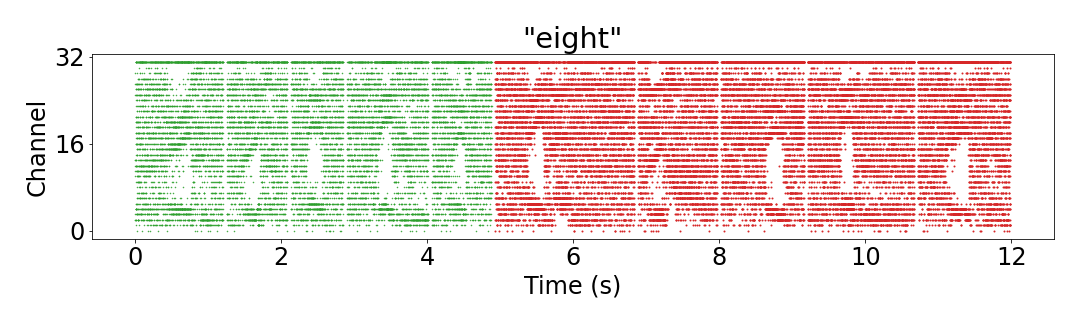}
    \end{subfigure}
    \caption{%
        \textbf{Examples of formant spike-data.}
        One spike is generated on each of the four channels that correspond to the most active auditory frequency bands in time bins of 1~ms.
        Ten subsequent samples are illustrated for each of the specified digits.
        Green color indicates samples for which true positive tests were generated by all single-neuron systems of Fig.~\ref{fig:single_neuron}, while red indicates false negatives.
    }
    \label{fig:data}
\end{figure*}


In order to simplify the input data, we extracted \textit{formants}---spectral maxima---from the vocal recordings as in \cite{coath2014robust}, rather than using them in full.
Formants are the characteristic properties that primarily distinguish vocal sounds from each other in human perception.
%
%
Here, we extracted the four ``first" formants---that is, the ones with the highest amplitudes---from the speech data out of 32 frequency bands using Sinewave speech analysis for MATLAB \cite{sinewave}.
This number of frequency bands was motivated by the analysis behind the design of a neuromorphic cochlea \cite{chan2007aer}.
The formants were converted into spike-trains by, in time-bins of 1~ms, generating one spike on the four input channels corresponding to the frequency bands of the formants---i.e.\ the instantaneous spectral maxima.
Examples of the resulting spike data are presented in \textbf{Fig.~\ref{fig:data}}.
%


After processing the data through the \gls{snn} layers according to \textbf{Fig.~\ref{fig:network}}, it was simplified by converting the resulting spike trains to spike counts for each presented data sample---that is, for each spoken-digit utterance.
A logistic regression model performing the one-vs.-rest classification task, implemented using the Python machine learning library scikit-learn, was then fitted to the spike-count data.

\subsection{The Time-Difference Encoder}
\label{sec:tde}

\begin{figure}[b!]
    \centering
    \includegraphics[width=\columnwidth]{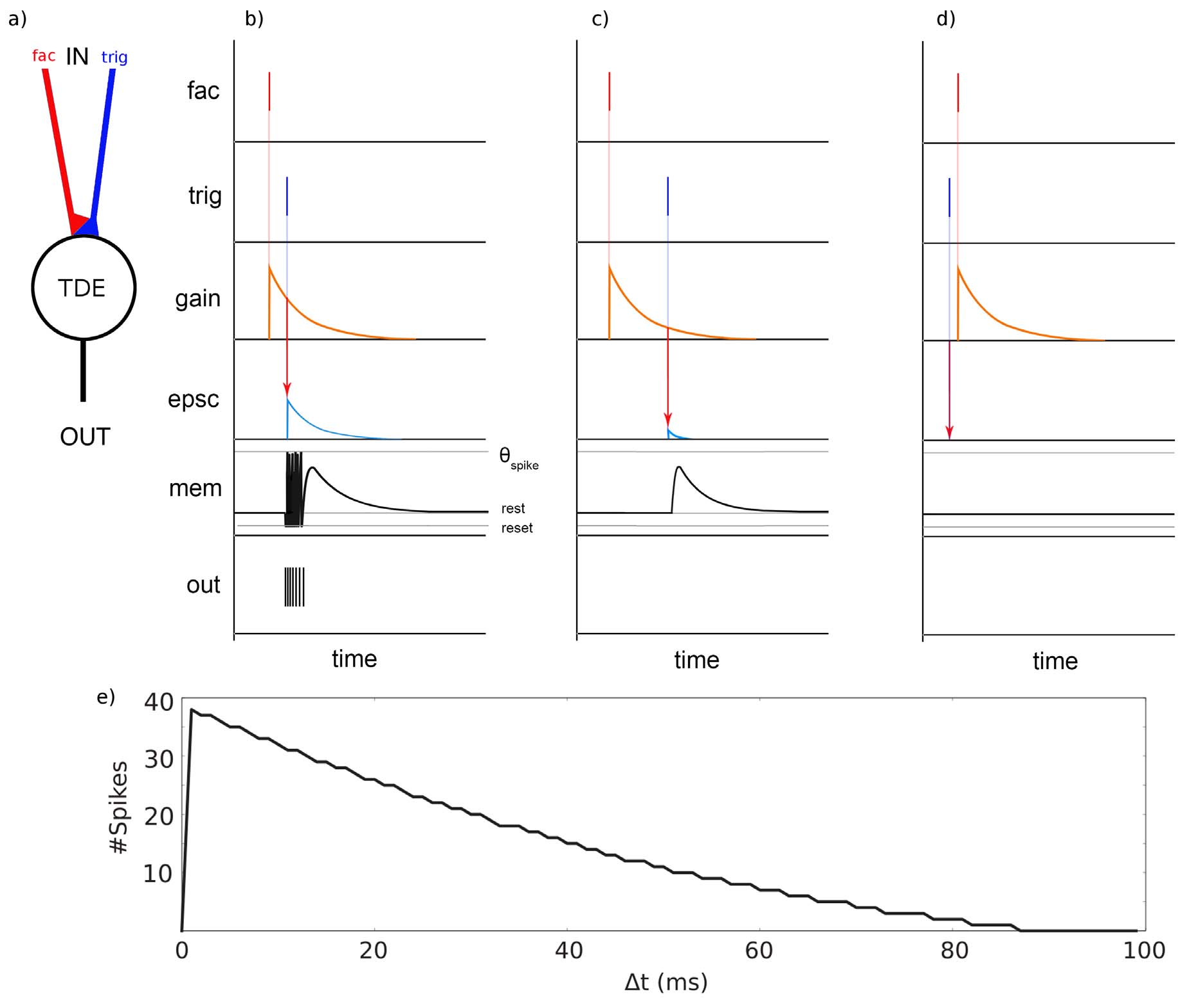}
    \caption{%
        \textbf{Basic principle of the time difference encoder (TDE).}
        Adapted from \cite{angelo2020motion} and \cite{gutierrez2022digital}.
        \textbf{(a)}
        A TDE neurosynaptic unit.
        Input received by the trigger (trig) synapse is gated by the facilitatory (fac) synapse with a gain that depends on the temporal difference between the two inputs.
        \textbf{(b)--(d)}
        TDE responses for small, large, and negative time differences, respectively.
        \textbf{(e)}
        TDE time-difference spike-response curve.
    }
    \label{fig:tde}
\end{figure}

The \glsentryfull{tde} \cite{milde2018elementary} is a computational element for encoding spike-time differences in \glspl{snn}, see \textbf{Fig.~\ref{fig:tde}}.
Each TDE unit consists of one current-based leaky integrate-and-fire (CuBa LIF) neuron with an extra ``gain" compartment and two input synapses: the facilitatory synapse and the trigger synapse.
Spikes received in the facilitatory synapse give rise to an exponentially decaying gain factor.
Spikes received in the trigger synapse generate an \gls{epsc}, which is integrated into the membrane potential, and which has an amplitude that is proportional to the instantaneous value of the gain factor.
Therefore, the current integrated by the neuron depends on the temporal difference between presynaptic spikes, which thereby becomes encoded in a burst of spikes.
Both the number of spikes and the inter-spike intervals depend on the time difference between the two inputs.
Thus, one \gls{tde} neuron is required for each pair of input channels---here, auditory frequency bands---for which a time difference is to be encoded.

\subsubsection{TDE Topology}

%
The aim of the TDE layer is to encode the temporal changes of the formant tracks, similarly to how, in image processing, convolutional layers are used to extract the basic features present in an image.
To do so, the TDEs should capture information about the temporal changes within single formant tracks.
Consequently, the activation of some TDEs triggered by spikes corresponding to different formant tracks needs to be avoided.
To this aim, a maximum lateral distance of $d_{\textrm{max}}$~=~3 between presynaptic input channels was defined.
As described in the previous section, each TDE neuron receives input from two different input channels.
Thus, setting up one \gls{tde} for each possible pairwise input combination results in 180 \gls{tde} neurons.

\subsubsection{TDE Implementation}

The \gls{tde} layer of the \gls{snn} architecture, see \textbf{Fig.~\ref{fig:network}}, was simulated using the Python package Nengo \cite{bekolay2014nengo}.
The parameters of the \glspl{tde} were set according to a grid search done using spoken digits from the Speech Commands dataset \cite{warden2018speech}, see \textbf{Table~\ref{tab:params}}.
Using a dataset with different speakers for tuning the feature extractors than for training the final classifier is similar to a development process in which the feature extractors are tuned during development, while the classifier is trained on the end-user of the consumer product.

\subsection{Excitatory--Inhibitory Disynaptic Elements}
\label{sec:ei_elements}

\begin{figure}[b!]
    \centering
    \begin{subfigure}[b]{0.7\columnwidth}
        \centering
        \includegraphics[width=\textwidth]{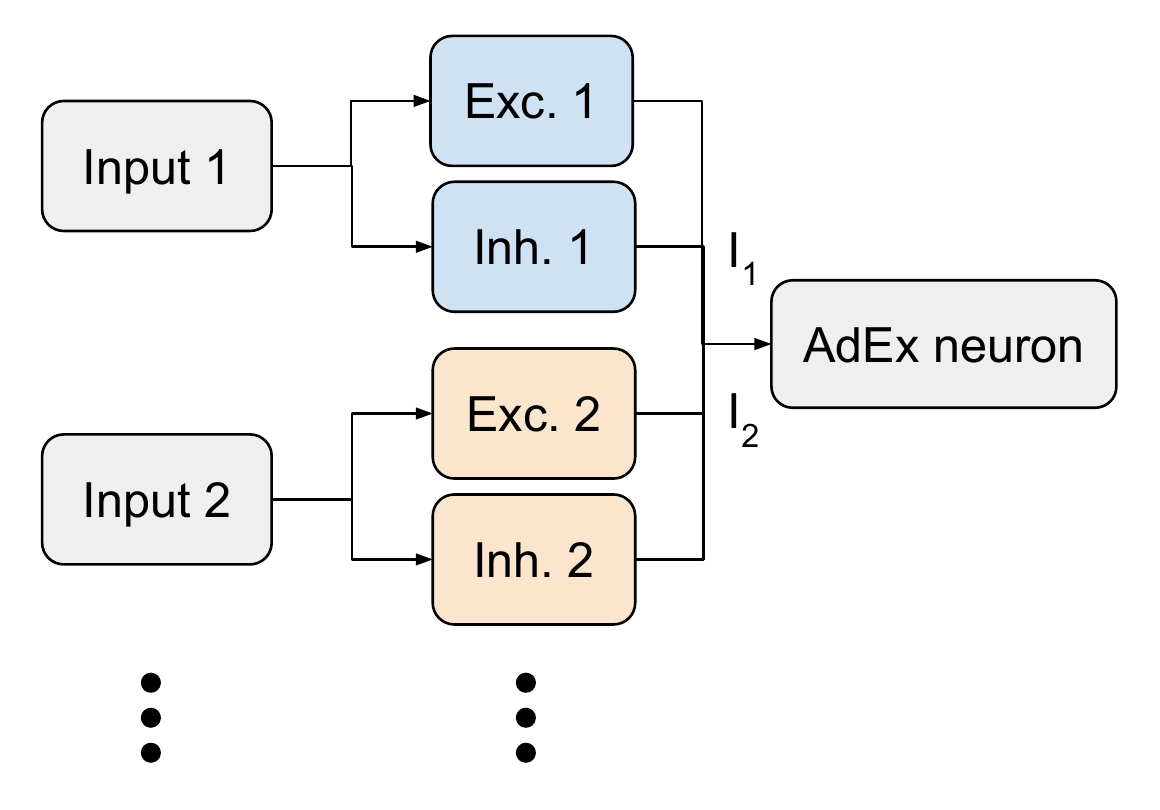}
        \caption{}
     \end{subfigure}
     \begin{subfigure}[b]{0.7\columnwidth}
        \centering
        \includegraphics[width=\textwidth]{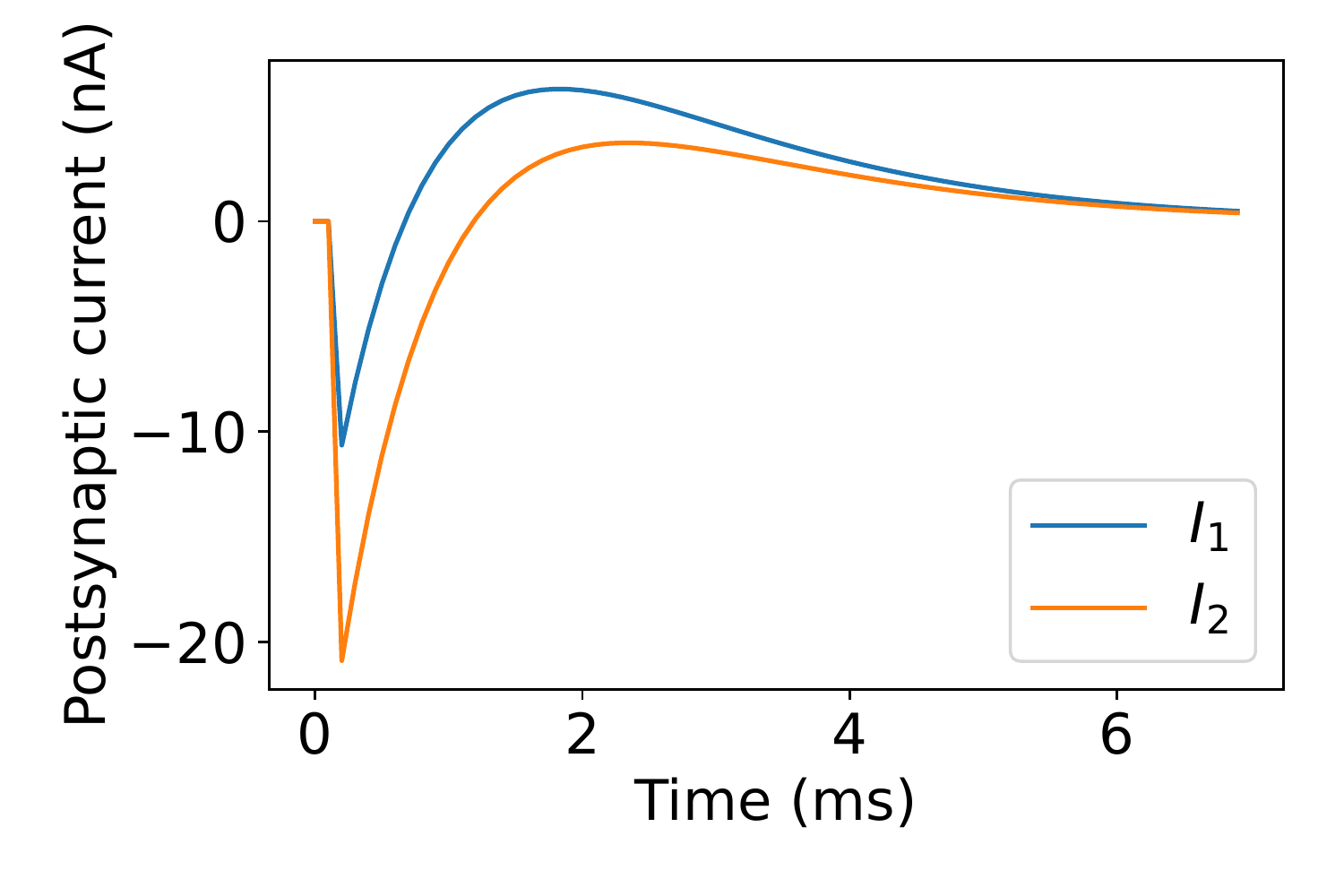}
        \caption{}
    \end{subfigure}
    \caption{%
        \textbf{Basic principle of disynaptic \glsentryfull{e-i} elements.}
        \textbf{(a)}
        One AdEx neuron receiving inputs through two different E--I elements, each consisting of one excitatory (Exc.)\ and one inhibitory (Inh.)\ dynamic synapse with different time constants.
        \textbf{(b)}
        Postsynaptic currents ($I_1$ and $I_2$) of the two E--I elements, which differ due to circuit inhomogeneity (``device mismatch"), resulting in different temporal delays and amplitudes of the postinhibitory excitations.
    }
    \label{fig:ei}
\end{figure}

The \glsentryfull{e-i} disynaptic elements \cite{sandin2020delays} use a summation of one excitatory and one inhibitory postsynaptic current for each presynaptic connection to generate an effectively delayed excitation of the neuronal membrane potential, see \textbf{Fig.~\ref{fig:ei}}.
Thereby, they imitate the neurobiological phenomenon of postinhibitory rebound.
The \gls{e-i} elements offer a low-resource alternative to dedicated delay mechanisms or dendritic emulation for \glspl{snn}, which are costly to implement in neuromorphic hardware.
Implementation of \gls{e-i} elements has previously been investigated in the DYNAP-SE neuromorphic processor \cite{sandin2020delays, nilsson2020integration, nilsson2022spatiotemporal}, where they were used to leverage the heterogeneity of the analog neuron and synapse circuits as a source of variability in generated delays.

\subsubsection{E--I Topology}

Based on the motivations described for the \gls{tde} topology, also the \gls{e-i} layers were set up with a maximum input distance of $d_{\textrm{max}}$~=~3.
Here, based on the topology of the \gls{dstc} network \cite{nilsson2022spatiotemporal}, we translated this lateral input range into each \gls{e-i} neuron being connected to four adjacently placed input channels---each via one \gls{e-i} element.
In the first \gls{e-i} layer, we set up one neuron for each continuous sequence of four adjacent input channels, which resulted in a total of 29 \gls{e-i} neurons.
To make the search space of \gls{e-i} neurons similar in size to that of the \glspl{tde}, we set up six duplicates of every \gls{e-i} neuron, resulting in a total of 174 neurons.


As each E--I neuron serves as a small spatiotemporal feature detector, we furthermore added a second, subsequent E--I layer, to investigate the effect of depth and, consequently, longer effective time constants in this architecture.
For the forward connections to the second \gls{e-i} layer, the aforementioned connection pattern was repeated for each of the six duplicates of the first layer, respectively, resulting in a total of 156 neurons.

\subsubsection{E--I Implementation}

The \gls{e-i} layers were simulated using the teili toolbox \cite{milde2018teili} for the Brian~2 \gls{snn} simulator \cite{stimberg2019brian2}.
Teili offers implementations of the analog neuron and synapse circuit-models of the DYNAP-SE \cite{moradi2018dynaps}---the \gls{adex} neuron model and \gls{dpi} synapse circuit---which were used in these simulations with the teili neuron and synapse models \texttt{DPI} and \texttt{DPISyn}, respectively.
The time-constants of the \gls{e-i} elements were set such that they generate an effective delay of roughly 2~ms---motivated by the delay magnitudes in \cite{coath2014robust} (original description in \cite{sheik2012emergent})---and the weights were set such that the \gls{e-i} neurons generate an output spike in the case of at least four temporally coincident input spikes, see \textbf{Table~\ref{tab:params}}.
All other parameters were kept in the default values of the teili neuron and synapse models.
As the topology of the \gls{e-i} layers described here relies on variance in the synapse dynamics to form different feature detectors, we used the device-mismatch functionality of teili to apply such variance to the synaptic time constants and weights of all \gls{e-i} synapses.
The mismatch follows a Gaussian distribution with a given standard deviation, here set to 50\% to strongly emphasize this effect---whereas the default value in teili is 20\%.

\begin{table}[tb]
    \centering
    \caption{%
        \textbf{Synaptic parameter values.}
    }
    \begin{tabular}{l l c c}
        \toprule
        \textbf{Model} & \textbf{Parameter}         & \textbf{Symbol}           & \textbf{Value} \\
        \midrule
        TDE     
                & Facilitatory time constant         & $\tau_{\textrm{fac}}$     & 8 ms      \\
                & Trigger time constant             & $\tau_{\textrm{trig}}$    & 2 ms      \\
                & Facilitatory weight                & $w_{\textrm{fac}}$        & 50,000    \\
                & Trigger weight                    & $w_{\textrm{trig}}$       & 50,000    \\
        \midrule
        E--I    
                & Excitatory time constant          & $\tau_{\textrm{e}}$       & 1.5 ms    \\
                & Inhibitory time constant          & $\tau_{\textrm{i}}$       & 1 ms      \\
                %
                %
                & Excitatory weight                 & $w_{\textrm{e}}$          & 105 nA    \\
                & Inhibitory weight                 & $w_{\textrm{i}}$          & -147 nA   \\
                & Time constant standard deviation  & $\sigma_{\tau}$           & 50\%     \\
                & Weight standard deviation         & $\sigma_{w}$              & 50\%     \\
        \bottomrule
    \end{tabular}
    \label{tab:params}
\end{table}

\section{Results}
\label{sec:results}

\subsection{Permutation Importance of Neurons}

\begin{figure}[tb]
    \centering
    \includegraphics[width=\columnwidth]{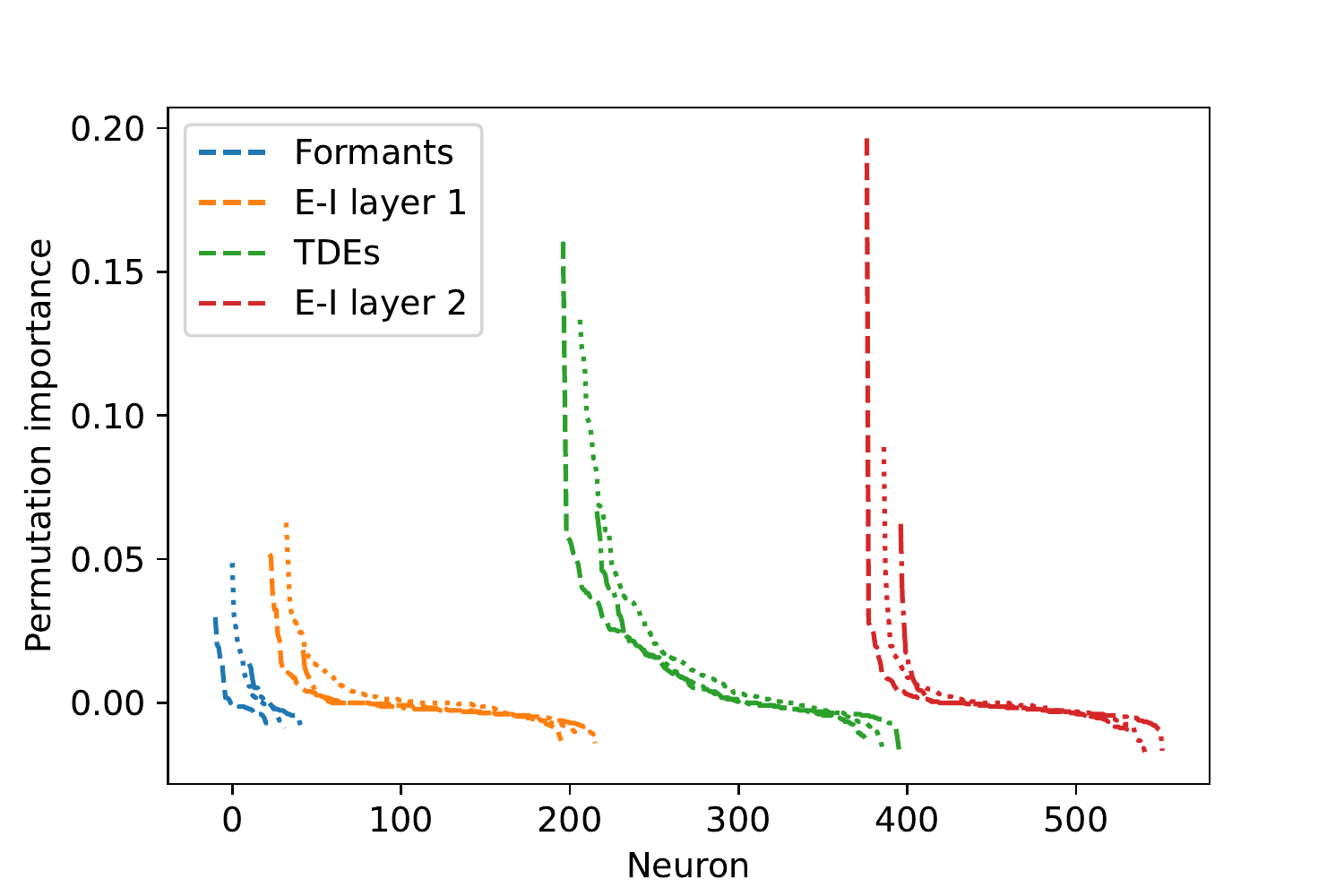}
    \caption{%
        \textbf{Permutation importance of neurons.}
        The different line styles correspond to the keywords ``one" (dashed), ``two" (dotted), and ``three" (dash-dotted), respectively, and the corresponding plots are offset with x~=~10 for improving visibility.
        The permutation importance was evaluated on the test set for a logistic regression model fitted on the training data from all layers of the neural network.
        The neurons are sorted internally within each layer by magnitude of permutation importance.
    }
    \label{fig:importance}
\end{figure}

\textbf{Fig.~\ref{fig:importance}} shows the permutation importance of all neurons, including the formant input channels. 
The permutation importance is defined as the decrease in the model score when the values of a single feature---in this case the individual spike-counts of each neuron---are randomly permuted \cite{breiman2001random}. 
This is computed by classifying the outputs of each layer in the network described in Fig. \ref{fig:network} with a single linear classifier.
Hence, the permutation importance gives an indication about the relative amount of information about a specific class that is encoded by each neuron.
In the case of the TDE layer, as well as E--I layer 2, the subset of encoders with the highest permutation importance have values 3 to 4 times higher than for the formants.
These results affirm the intuition that these computational elements can be useful for the extraction of information about spatiotemporal patterns in the input data.

\begin{figure}[b]
    \centering
    \includegraphics[width=\columnwidth]{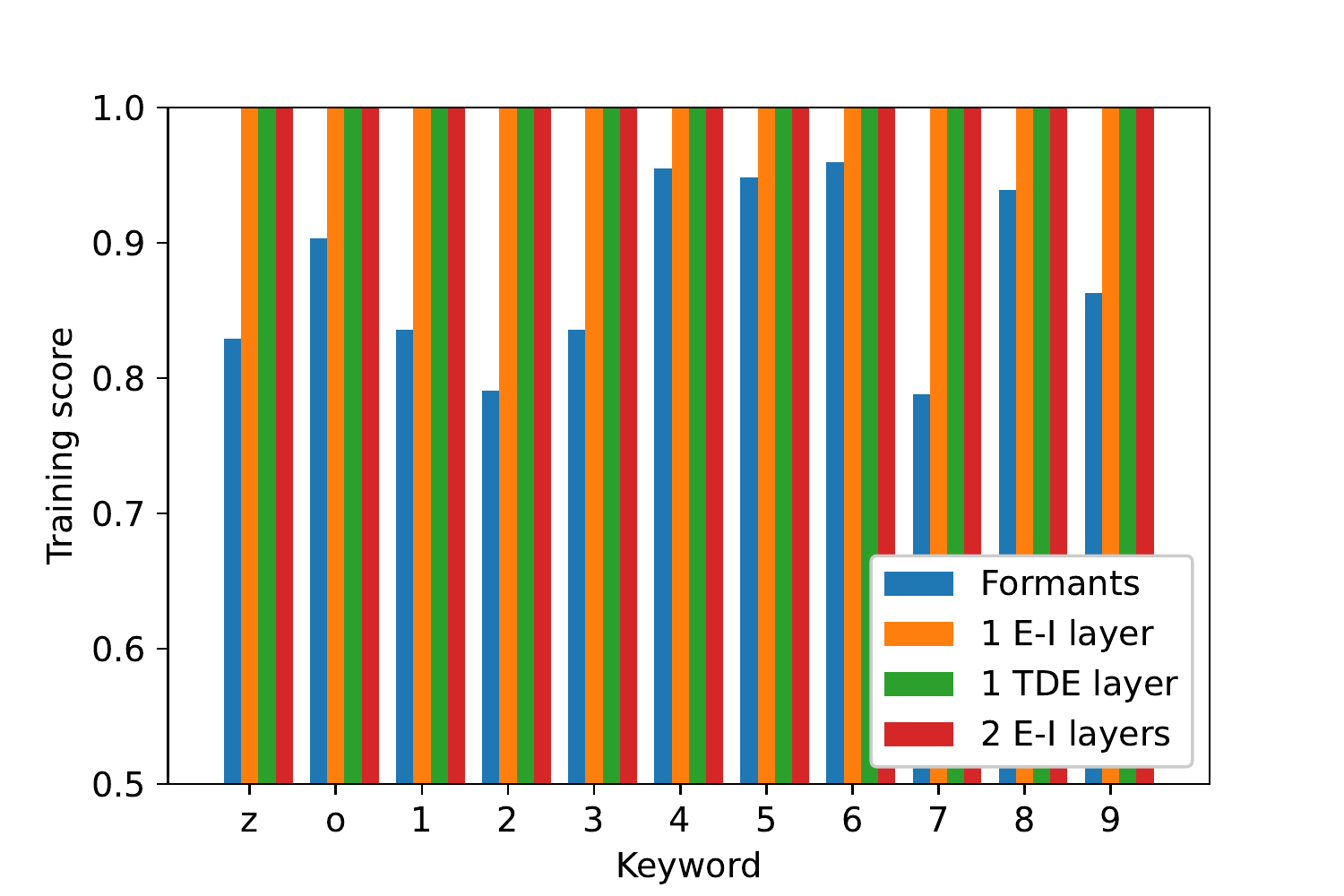}
    \caption{%
        \textbf{Mean accuracy on the training set using all neurons of the indicated architecture.}
        In this case, the linear classifiers receive spike-counts from the last layer of the architecture as well as from the preceding layers via bypass connections.
        The total numbers of neurons used in the different architectures are thus:
        Formants: 32;
        1 E--I layer: 32+174;
        1 TDE layer: 32+180;
        2 E--I layers: 32+174+156.
    }
    \label{fig:scores_train}
\end{figure}

\subsection{Linear Classification}

\textbf{Fig.~\ref{fig:scores_train}} shows the mean accuracy on the linear classification of the training dataset fed through the different layers of the network.
Each classifier receives spike-counts from the indicated layer, as well as from the layers preceding it.
%
%
These results show that all the investigated encoders extract sufficient information for a complete linear classification of the training data (444 samples in total, 100\% accuracy).
Furthermore, the results indicate that the extracted formants, without further processing, provide sufficient information for a fairly accurate linear classification, ranging between 79\% and 96\% average accuracy for different keywords, whereas random guessing would give 50\%.

\subsection{Classification with Few Neurons}

\begin{figure}[tb]
     \centering
     \begin{subfigure}[b]{\columnwidth}
        \centering
        \includegraphics[width=\textwidth]{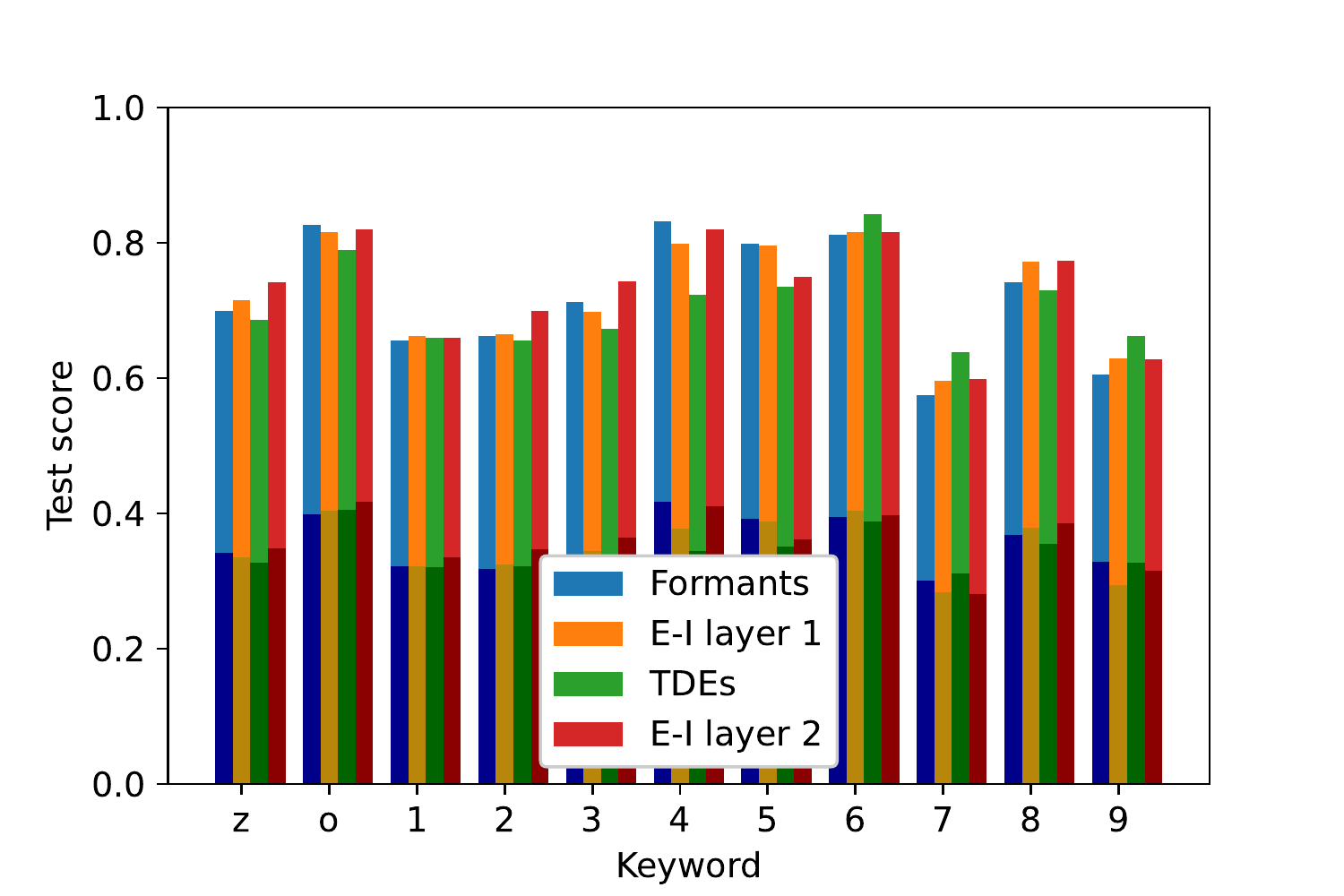}
        \caption{}
        \label{fig:score_single}
    \end{subfigure}
    \begin{subfigure}[b]{\columnwidth}
        \centering
        \includegraphics[width=\textwidth]{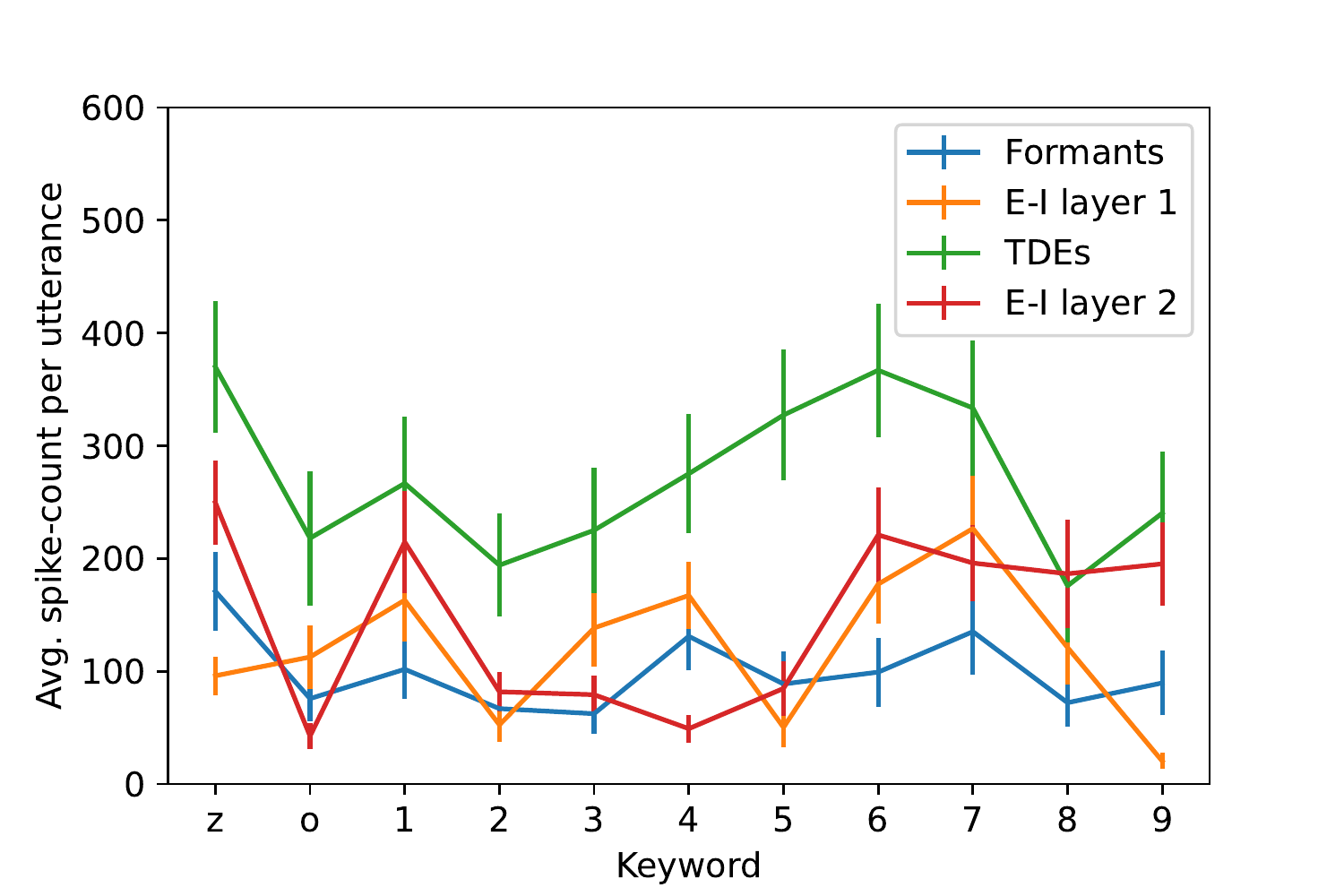}
        \caption{}
        \label{fig:mean_counts}
    \end{subfigure}
    \caption{%
        \textbf{Classification with single neurons.}
        \textbf{(a)}
        Mean accuracy on the test set.
        Light and dark colors denote the share of true positives and negatives, respectively, in the total number of true classifications.
        \textbf{(b)}
        Mean number of spikes per keyword utterance.
        The error bars denote the standard deviation.
    }
    \label{fig:single_neuron}
\end{figure}

\textbf{Fig.~\ref{fig:score_single}} shows the mean accuracy in the classification of the testing dataset with single neurons.
The neurons were selected by highest accuracy on the training set.
There is no clear improvement in accuracy when using the outputs from single elements of the \gls{tde} and \gls{e-i} layers compared to single formant channels.
Moreover, there is no systematic improvement from adding the second E--I layer.
\textbf{Fig.~\ref{fig:score_single}}, furthermore, illustrates the share of true positives and true negatives, respectively, in the total number of true classifications.
This share is fairly evenly divided and, therefore, does not indicate that classification errors would be concentrated to positive or negative predictions.
\textbf{Fig.~\ref{fig:mean_counts}} shows the mean number of spikes that each of the neurons generate for processing a single utterance of the different keywords.
While the spike-counts of both the \glspl{tde} and the \gls{e-i} neurons may be subject to optimization by parameter tuning, it is interesting to note that the \glspl{tde} generate in the order of 100 spikes more than the other neurons for almost all keywords.
This is an important metric since spike generation and routing makes out the majority of dynamic power usage, and thus a substantial part of total power usage, of a neuromorphic processor such as the DYNAP-SE \cite{moradi2018dynaps}.

\begin{figure*}[tb]
    \centering
    \begin{subfigure}[b]{0.49\textwidth}
        \centering
        \includegraphics[width=\textwidth]{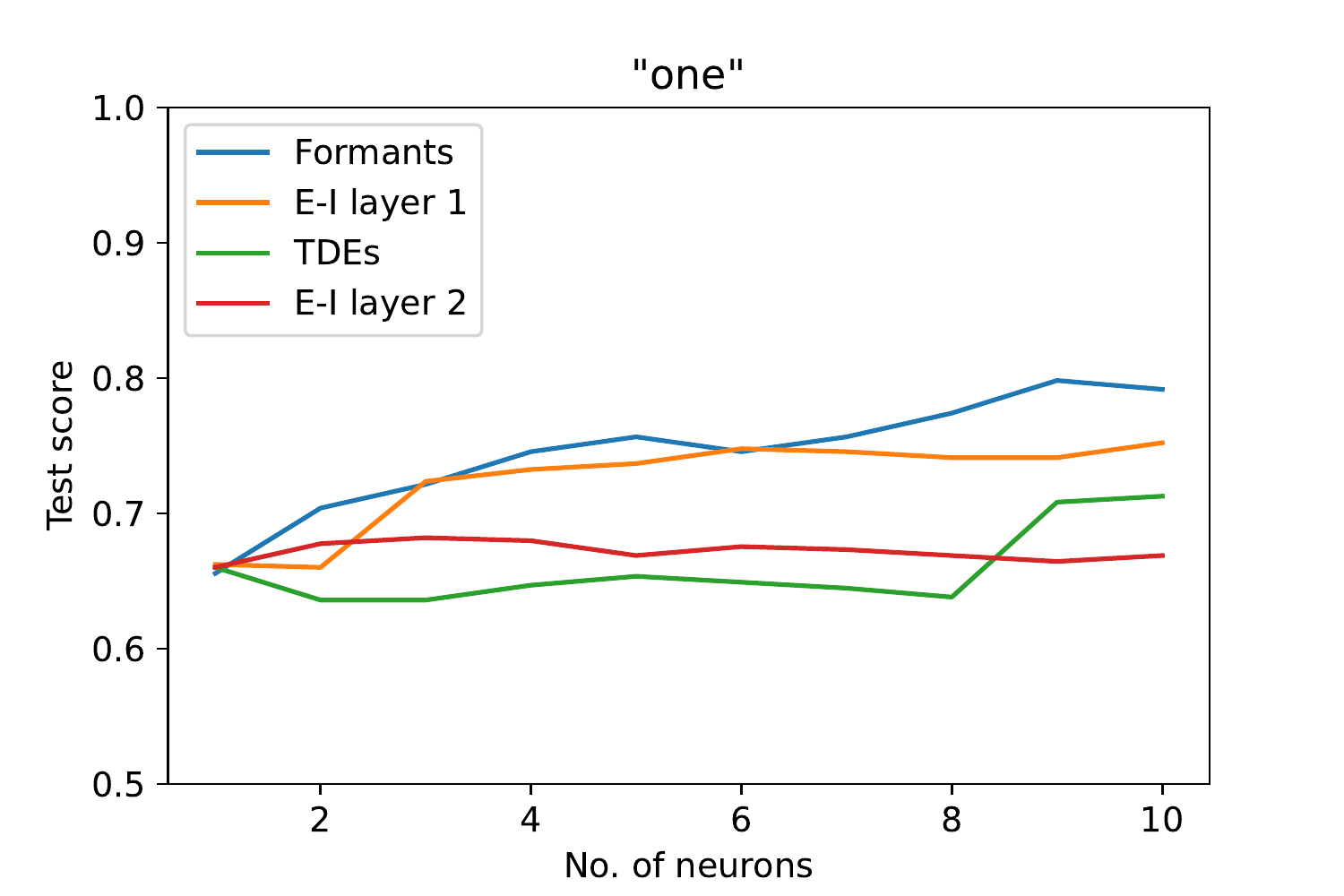}
    \end{subfigure}
    \begin{subfigure}[b]{0.49\textwidth}
        \centering
        \includegraphics[width=\textwidth]{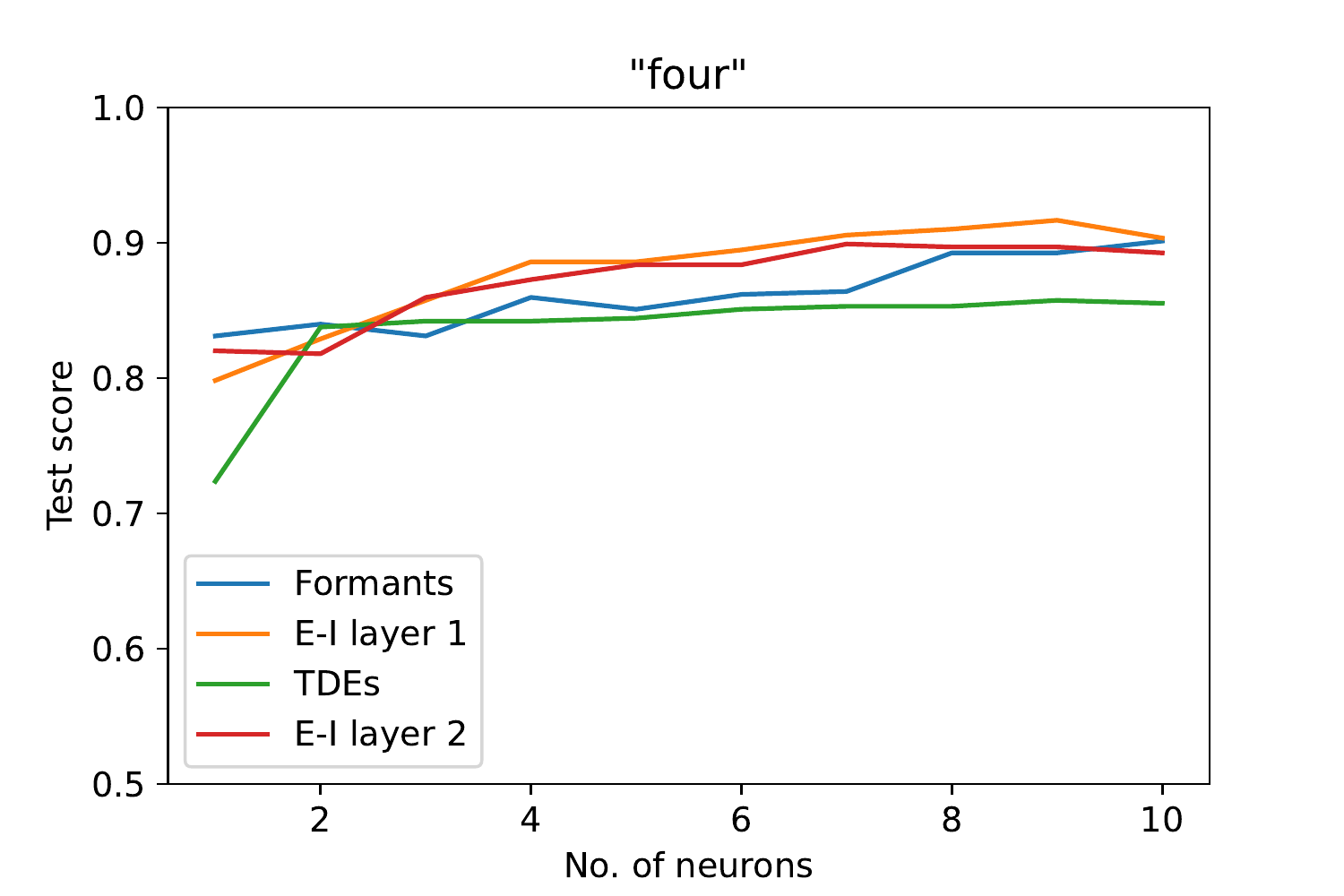}
    \end{subfigure}
    \\
    \begin{subfigure}[b]{0.49\textwidth}
        \centering
        \includegraphics[width=\textwidth]{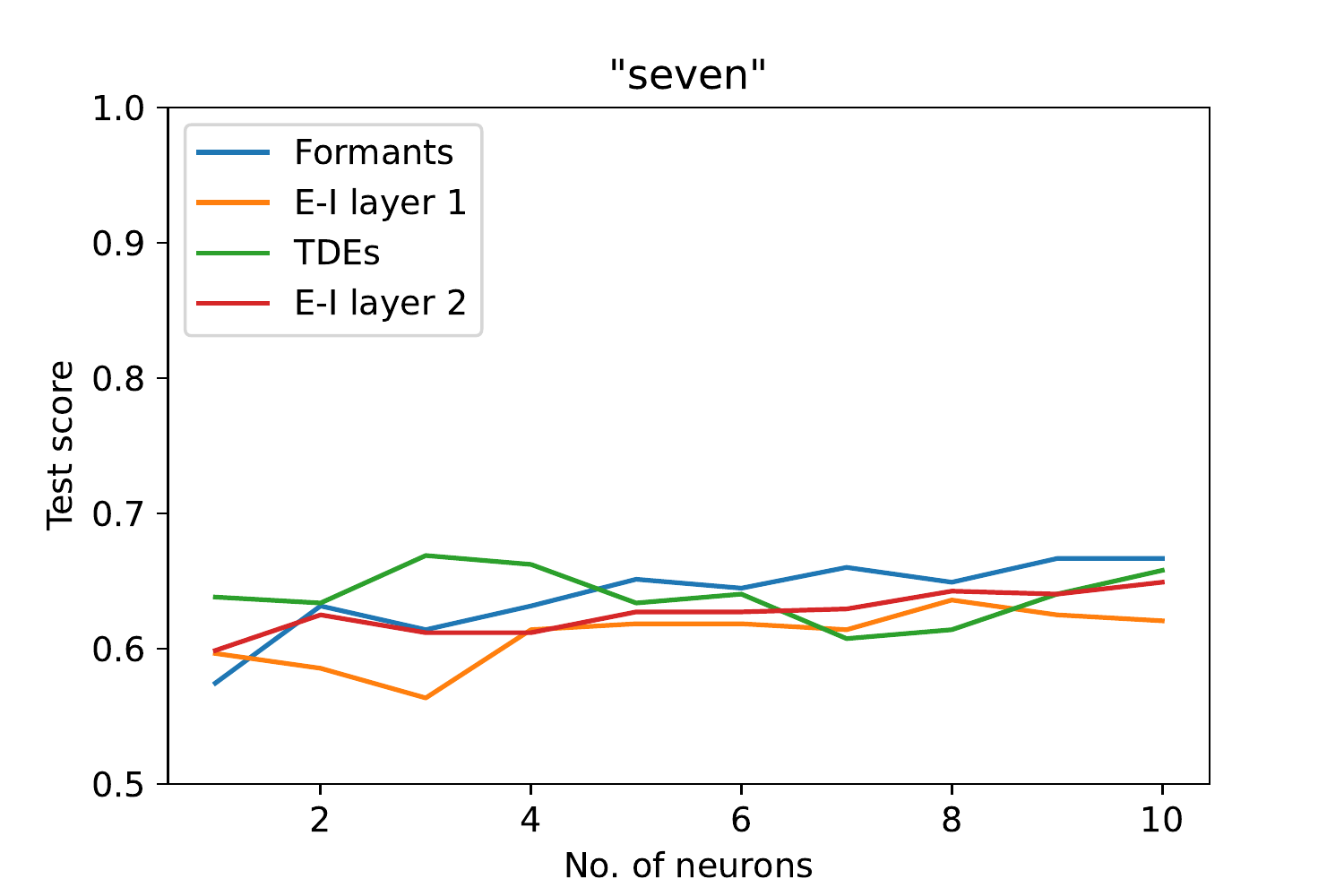}
    \end{subfigure}
    \begin{subfigure}[b]{0.49\textwidth}
        \centering
        \includegraphics[width=\textwidth]{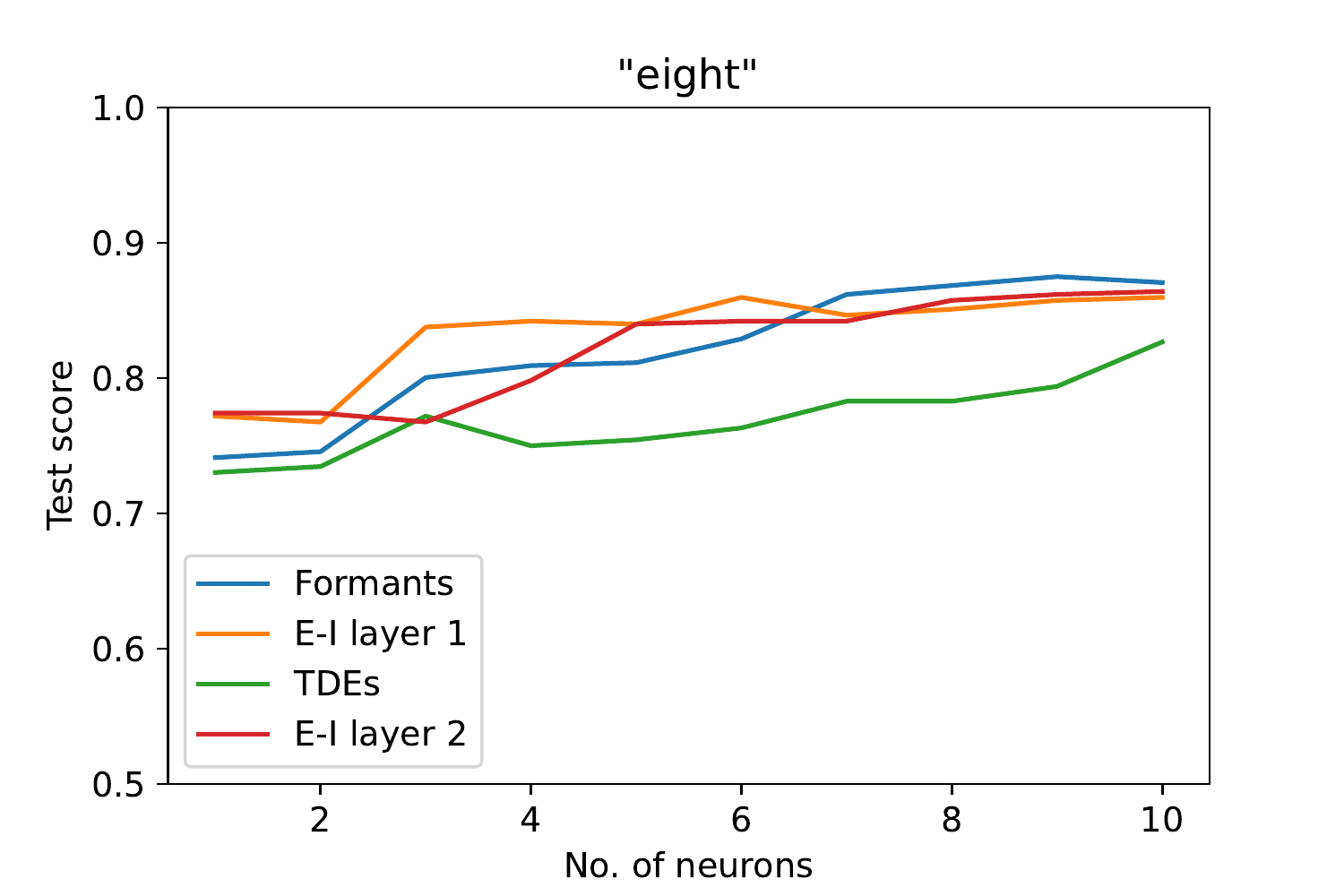}
    \end{subfigure}
    \caption{%
        \textbf{Mean accuracy on the test set using limited numbers of read-out neurons.}
        The first neuron was selected by maximum training score, and the subsequent neurons were selected by maximum permutation importance.
    }
    \label{fig:scores_few}
\end{figure*}

\textbf{Fig.~\ref{fig:scores_few}} shows the mean accuracy of the different network layers on four different keywords in the test set when increasing the number of neurons of \textbf{Fig.~\ref{fig:single_neuron}} to up to 10.
For each layer, the first neuron was, as previously, selected by highest training score, while the subsequent neurons were selected by highest permutation importance.
Selecting a subset of neurons that show high permutation importance for the classification slightly improves the results from using single neurons.
On the other hand, there is no clear improvement from using TDEs or E-I elements instead of using a subset of the formant channels.
%

The results presented in this section highlight a lack of generalization of the system.
However, an interesting comparison is that, in another work on keyword spotting with \glspl{snn} on the same dataset, use of 4, 16, and 36 neurons in the feature extraction layer resulted in classification accuracies of about 9\%, 27\%, and 61\%, respectively \cite{wu2018biologically}.

In the case of the TDEs, the time constants and weights of all units are set to the same values.
Similarly, the parameters of the E--I elements have values drawn from Gaussian distributions with constant mean values. 
In order to create sensitivity to the characteristic spatiotemporal features of each keyword, effectively improving the classification of unknown samples, inhomogeneous distributions of parameter values are likely required for both encoders.
Furthermore, as it is not feasible to tune hundreds of parameters manually or in a grid search, learning would be required to optimize such parameter values \cite{perez2021neural, bouanane2022impact}.
In the case of the E--I elements, this would mean optimizing the choice of inhomogeneous neuron and synapse circuits for each E--I neuron, rather than, like here, relying on sampling from a large number of randomly inhomogeneous neurosynaptic units.

\section{Conclusion}
\label{sec:conclusion}

We have presented results from a study comparing the use of two different neurocomputational elements for spatiotemporal encoding---the \gls{tde} and disynaptic \gls{e-i} elements---for resource-efficient keyword spotting feasible for implementation in always-on neuromorphic hardware.
While our results on the training data show that both of these encoders enable a complete one-vs.-rest linear classification, the results on the test data show no clear improvement compared to direct linear classification of the formant data.
%
%
%
%
%
%
%
%
%
A major source for improvement in the use of both encoders is likely found in using heterogeneous weights and time constants, and in optimizing these with training to maximize detection accuracy on specific keywords.
%
%



\section*{Acknowledgments}
This work was partially funded by
The Kempe Foundations under Contract~JCK-1809,
ECSEL JU under Grant No.~737459,
the CogniGron research center,
and the Ubbo Emmius Funds (University of Groningen).

\bibliographystyle{IEEEtran}
\bibliography{references}

\end{document}